\documentclass[10pt, journal,compsoc]{IEEEtran}
\IEEEoverridecommandlockouts
\usepackage{cite}
\usepackage{amsmath,amssymb,amsfonts}
\usepackage{algorithmic}
\usepackage{graphicx}
\usepackage{subcaption}
\usepackage{textcomp}
\usepackage{xcolor}
\usepackage{amsthm}
\usepackage{amsmath}
\theoremstyle{definition}
\newtheorem{defn}{\textit{Definition}} 
\newcommand{\threestage}{Three Stages Recurrent Temporal Learning Framework}

\newcommand{\todo}[1]{\textcolor{red}{(#1)}}
\newcommand{\matr}[1]{\mathbf{#1}}
\newcommand{\vecr}[1]{\mathbf{#1}}

\newcommand{\comment}[1]{}
\def\BibTeX{{\rm B\kern-.05em{\sc i\kern-.025em b}\kern-.08em
    T\kern-.1667em\lower.7ex\hbox{E}\kern-.125emX}}
\begin{document}

\title{Encoder-Decoder Architecture for Supervised Dynamic Graph Learning: A Survey}

\author{\IEEEauthorblockN{Yue Cai Zhu\IEEEauthorrefmark{1}, Fuyuan Lyu\IEEEauthorrefmark{2}, Chengming Hu\IEEEauthorrefmark{3}, Xi Chen\IEEEauthorrefmark{4}, Xue Liu\IEEEauthorrefmark{5}}\\
\IEEEauthorblockA{\textit{School of Computer Science} \\
\textit{McGill University}\\
Montreal, Canada \\
Email:
\IEEEauthorrefmark{1}yue.c.zhu@mail.mcgill.ca, \IEEEauthorrefmark{2}fuyuan.lyu@mail.mcgill.ca, \IEEEauthorrefmark{3}chengming.hu@mail.mcgill.ca, \IEEEauthorrefmark{4}xi.chen11@mail.mcgill.ca, \IEEEauthorrefmark{5}xueliu@cs.mcgill.ca}
}

\maketitle

\begin{abstract}

In recent years, the prevalent online services generate a sheer volume of user activity data.
Service providers collect these data in order to perform client behavior analysis, and offer better and more customized services.
Majority of these data can be modeled and stored as graph, such as the social graph in Facebook,
user-video interaction graph in Youtube.
These graphs need to evolve over time to capture the dynamics in the real world, leading to the invention of dynamic graphs.
However, the temporal information embedded in the dynamic graphs brings new challenges in analyzing and deploying them.
Events staleness, temporal information learning and explicit time dimension usage are some example challenges in dynamic graph learning.

In order to offer a convenient reference to both the industry and academia, this survey presents the \threestage~based on dynamic graph evolution theories, so as to interpret the learning of temporal information with a generalized framework.
Under this framework, this survey categories and reviews different intelligent encoder-decoder architectures for supervised dynamic graph learning.
We believe that this survey could supply useful guidelines to researchers and engineers in finding suitable graph structures for their dynamic learning tasks.
\end{abstract}


\section{Introduction}
In the data explosion era, the amount of data increases exponentially. 
Most of data can be viewed as a graph. 
Graph is a data structure that consists of nodes and edges.
It is designed to model and store data that contains not only features for different entities, but also relations between them.
Graph analysis has long been an important research topic.
Previous works \cite{Gilmer2017, velivckovic2017graph, wu2020comprehensive} assume that the underling graph is a static graph which does not change over time.
But in real world, the entities modeled as graph present different temporal dynamic in node features and relations.
Dynamic graph is developed to model and store such an evolving graph.
The extra time dimension brings temporal information to the graph's representation and reveals the causality embedded in its network dynamic\cite{masuda2020guide}.
However, such temporal information also increases the difficulty of analyzing graphs.

In recent year, utilizing machine learning techniques to analyze dynamic graph has become an emerging research topic\cite{sankar2020dysat, xu2020inductive,trivedi2017know}. 
Moreover, the prevalent online services generate a sheer volume of relational data, transaction data and interaction data.
They are modeled and stored as attributed dynamic graphs, such as social graph in Facebook \cite{weaver2013facebook} and user video interaction graph in Youtube\cite{benevenuto2008understanding}.
Those dynamic graph databases with rich attributes make supervised dynamic graph learning feasible and urge the industry to look for effective supervised dynamic graph learning methods\cite{rossi2020temporal,trivedi2017know, li2017attributed} . 
Therefore, we believe a survey of such methods is extremely helpful for the industry and the research community to exploit the potential of those databases.

There are multiple well-written surveys to summarize dynamic graph representation learning algorithms.
Kazemi et al.\cite{kazemi2020representation} focus on a broad topics of dynamic graph representation learning. 
Skarding et al.\cite{skardinga2021foundations} specialize in Graph Neural Network models for dynamic graph. 
They all follow the encoder-decoder learning framework proposed by Hamilton et al.\cite{hamilton2017representation}. With this framework, encoder generates graph embedding at node level, and decoder use the embedding to perform prediction/classification. 
Practitioners could assemble different encoder and decoder combinations to best fit their machine learning task. 
Moreover, the decoder can be modified to perform all dynamic graph learning tasks introduced in later sections. 
However, the aforementioned surveys do not focus on supervised learning methods, and they do not discuss how temporal information is learnt.

Our work is different from the previous ones mainly in that we develop the \threestage~based on dynamic graph evolution theories.
We use this framework to explain how dynamic graphs evolve over time and how different algorithms can learn the temporal information. 
It also gives a general form of these algorithms.

In the development of the mentioned framework, we found that using time as an input feature enables learning algorithms to
recognize temporal periodicity and vector clock\cite{lamport2019time}.
Vector clock is recently introduced to describe the phenomena that message sent from neighboring nodes to the target node requires different traversing time depending on their connection pattern.
This motivates us to categorize different algorithms by whether time is learnt implicitly or explicitly, namely Implicit Time and Explicit Time Learning Algorithm.   
Only Explicit Time learning Algorithms are capable to perform time prediction tasks which predict when a given graph updating event would happen. 
Time prediction task is recently recognized as one of the goals of dynamic graph learning.\cite{trivedi2017know, boschee2015integrated}

As a summary, we make the following contributions in this survey paper:
\begin{itemize}
    \item The \threestage~for dynamic graph learning. Under this framework, we discuss how temporal information is learnt by different dynamic graph learning algorithms.
    \item A list of different goals of dynamic graph learning which includes time prediction
    \item A review of the recent development of supervised dynamic graph learning for Discrete Time Dynamic Graph and Continuous Time Dynamic Graph
    \item Some interesting future research directions in dynamic graph learning according to the topics discussed
\end{itemize}

The organization of this survey is as follows: Sec.~\ref{ML-dg} introduces some background knowledge of machine learning in dynamic graph; Sec.~\ref{taxonomy} describes the taxonomy in this survey, mainly on the categories of dynamic graph, the encoder-decoder learning framework and the motivation of implicit/explicit time learning models categorization. Fig.~\ref{taxonomy} illustrates the taxonomy in this survey; in Sec.~ \ref{time-model}, we introduce the \threestage. With this framework we discuss how temporal information is learnt;  
We will then start the review from algorithms designed for \emph{Discrete Time Dynamic Graph (DTDG)} at Sec.~ \ref{sec-DTDG}, and then the algorithms designed for \emph{Continuous Time Dynamic Graph (CTDG)} at Sec.~ \ref{sec-CTDG}. 
Potential future directions are discussed in Sec.~ \ref{sec-fu}.
In Appendix, table summarizes the notations and abbreviations used in this work.
\begin{figure*}
\centering
\includegraphics[width=0.9\textwidth]{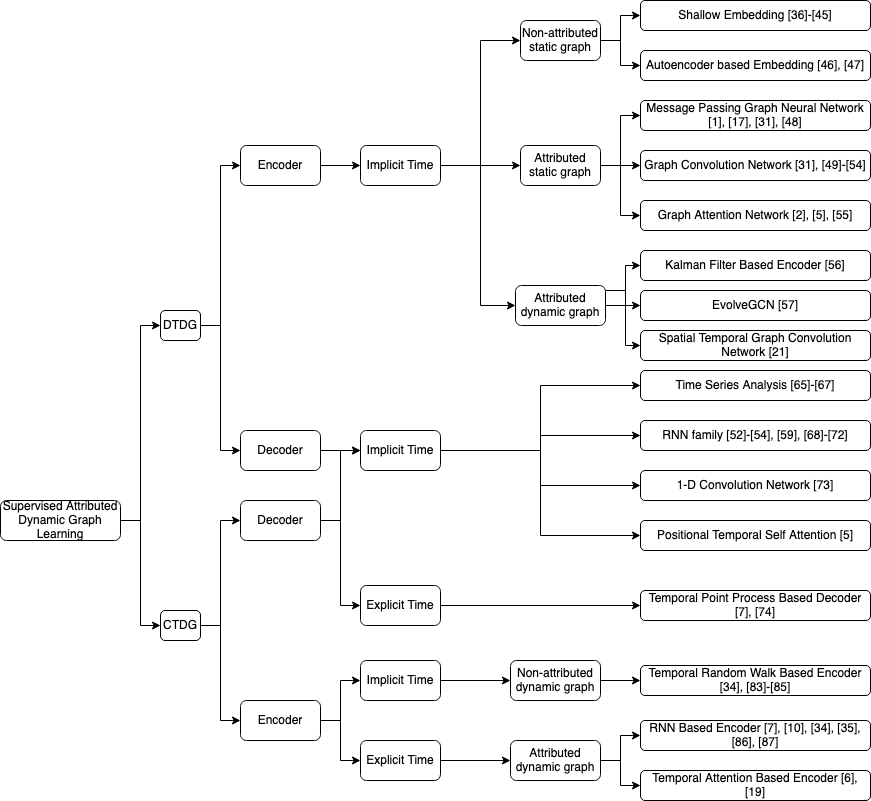}
\caption{Taxonomy}
\label{overview}
\end{figure*}

\section{Machine Learning In Dynamic Graph}
\label{ML-dg}
\subsection{Supervised Learning in Graph}
Supervised machine learning typically trains a machine learning model with historical data and conducts prediction or classification with the trained model during inference time.
In order to perform supervised training, the ground truth must be available in the historical data which is referred as label by convention.
Supervised learning in graphs differs from traditional machine learning in that it could be further categorized as node focus task, edge focus task and graph focus task\cite{scarselli2008graph}. In the following section, we use classification task as examples. Such paradigm can be easily extended to regression tasks. 

We can denote a graph as $G=(V,E,\matr{X})$ with $V=\{v_1, v_2, \dots, v_i, \dots, v_m\}$ as the set of nodes in $G$ where  $v_i \in G \wedge i \in [1,|V|]$, and $E=\{ e_{i,j}\}$ as the set of edges in $G$ where $e_{i,j}=(v_i,v_j,f_{i,j})$, $v_i, v_j \in V$ and $f_{i,j}$ represents the feature of edge $e_{i,j}$. 
$X$ is the node feature matrix with each row vector $x_{v_i} \in X$ stores the features for node $v_i \in V$.
Namely, there exists a one to one mapping between $V$ and the row vectors in $X$.
$\matr{X} \in \mathcal{R}^{|V| \times d}$ and $d$ is the number of features for a given node.

Then these three tasks can be defined as:
\begin{defn} \textit{
Node Focus Task: given a one to one mapping between $V_{\text{known}} = \{v_1, v_2, \cdots, v_i \}$ with $V_{\text{known}} \subset V$ and the label set $Y_{\text{known}} = \{y_1, y_2, \cdots , y_i\}$. The learning purpose is to predict $y_k$ for $v_k \in V_{\text{unknown}}$ where $V_{\text{unknown}} \cap V_{\text{known}} = \phi$, $V_{\text{unknown}} \subseteq V - V_{\text{known}}$.}
\end{defn}

\begin{figure*}[ht]
\begin{subfigure}{.33\textwidth}
  \centering
  \includegraphics[width=0.85\linewidth]{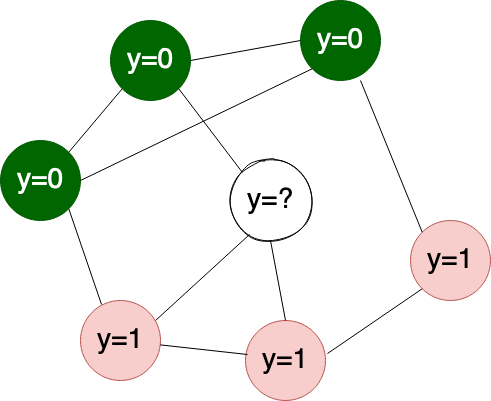}
  \caption{Node Focus Task}
    \label{n_focus}
\end{subfigure}
\begin{subfigure}{.33\textwidth}

\centering
\includegraphics[width=0.85\linewidth]{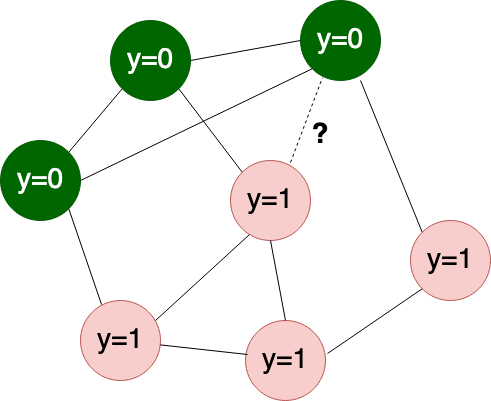}
\caption{Edge Focus Task}
\label{e_focus}
\end{subfigure}
\begin{subfigure}{.33\textwidth}

\centering
\includegraphics[width=0.87\linewidth]{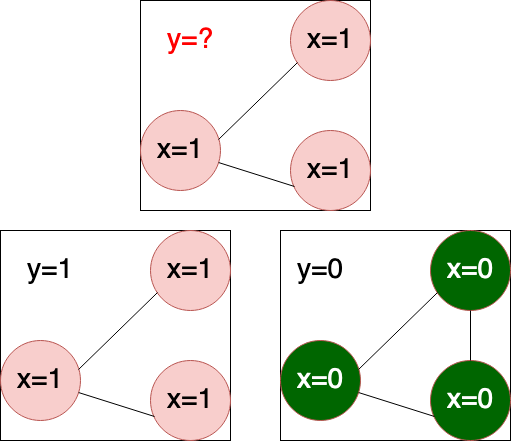}
\caption{Graph Focus Task}
\label{g_focus}
\end{subfigure}
\caption{Different Graph Learning Tasks}
\end{figure*}

Fig.~\ref{n_focus} illustrates the idea of node focus task. 
The color represents the node attribute. And each node has its label $y$.
Node focus task is to predict the unknown node labels.
As an example of node focus task, in community detection, the label assigned to each node would be the name of its associated community. 
If two nodes have the same label, they belong to the same community. 
It is also straightforward to see that node classification is node focus task.

\begin{defn} \textit{
Edge Focus Task: given a one to one mapping between $E_{\text{known}} \subset E$ and their corresponding labels $Y_{\text{known}}$. The learning purpose is to predict $y_{i,j}$ for $e_{i,j} \in E_{\text{unknown}}$ where $E_{\text{unknown}} = E-E_{\text{known}}$}
\end{defn}
As shown in Fig.~\ref{e_focus}, given a graph with node attribute and observed edges, edge focus task is to predict whether an edge exists between two given nodes.

\begin{defn} \textit{
Graph Focus Task: given a collection of graphs or sub-graphs $\{G_i\}$ and a one to one mapping between $\{G_i\}$ and the label set $\{y_i\}$, the learning purpose is to predict $y_{j}$ for $G_{j} \notin \{G_i\}$.}
\end{defn}

Given multiple graphs as in Fig.~\ref{g_focus}, some of them has known labels but some not.
Graph focus task is to predict the unknown graph label.

To sum up, the label $y$ to be predicted could be a cell in the adjacency matrix $A$ or an edge feature in edge focus tasks, one particular attribute in node attributes matrix $X$ for node focus tasks, or an numerical interpretation of the state tuple for graph focus tasks.

\subsection{Extrapolation and Interpolation Learning in Dynamic Graph}
In real world applications, we are facing the challenge of learning the network dynamic, namely the repetitive pattern in a dynamic graph's evolution over time.
Dynamic graphs have one more dimension than static graphs have, which is time.
Time dimension is usually stored as the timestamp when the observation of the graph or its components happens. 

Let's denote a dynamic graph as $G_T=O_T$, where $T=[t_1: t_n]$ is the time span from $t_1$ to $t_n$ and is referred as the observation period. 
$O_T = \{o_{t_1},o_{t_2},\cdots, o_{t_n}\}$ is the set of observations which are performed within the observation period $T$.
The observation could be a snapshot of graph $G_t = (V_t,E_t,\matr{X}_t)$ where $V_t$,$E_t$ and $\matr{X}_t$ are the snapshot of the nodes, edges and node features at time $t$,  a single node updating event $o_t=v_{i,t}$ or edge updating event $o_t=e_{\{i,j\},t}$ at time $t$.
Supervised learning in dynamic graphs could be extrapolation learning, interpolation learning or time prediction.

In \emph{Extrapolation Learning}, the purpose is to predict the label $Y_{t_{n+1}}$ at a future time based on the previous observations of a particular dynamic graph $G_T$ and ground truth labels $Y_T$ with the observation period $T=[t_1: t_n]$.
While in \emph{interpolation Learning}, the objective is to estimate the missing labels $Y_{t_i}$ such that $t_i \in T$ and $Y_{t_i} \notin Y_T$.
Extrapolation task predicts the future based on historical data while interpolation task gives estimation for the past.
Therefore, interpolation task is mainly used in data imputation. 
As an example, in stock market analysis, if the purpose is to predict the future trend, then it is extrapolation learning; if the purpose is to fill some missing data, then it is interpolation learning.

In \emph{Time Prediction}, the goal is to predict the time for a given incoming event. 
For example, predicting when the next crisis event would happen in \emph{Integrated Crisis Early Warning System (ICEWS)} \cite{trivedi2017know, boschee2015integrated}.
The different learning tasks for supervised dynamic graph learning are summarized in Table \ref{tab:learning-task}

	\begin{table*}
		\centering
		\caption{Supervised Learning task in Dynamic Graph}
		\label{tab:learning-task}
		\begin{tabular}{c|c|c|c} 
		\hline
			 & Extrapolation & Interpolation & Time Prediction \\
			\hline
			Node Focus & Predict the target node's attribute  & Estimate the target node's missing attribute  & Predict when a given node updating \\
			  & in the future & in the past & event will happen \\
			  \hline
			Edge Focus & Predict the target edge's status & Estimate the target edge's status &  Predict when a given link will be \\
				  & in the future & in the past & added or deleted \\
             \hline
			Graph Focus & Predict the given dynamic graph's attribute & Estimate the given graph's missing attribute &  Predict when the targeted graph will reach \\
						  & in the future & in the past & to a given state \\
			\hline
		\end{tabular}
	\end{table*}

\section{Taxonomy}
\label{taxonomy}
\subsection{Dynamic Graph Storage Model}
A dynamic graph represents a graph evolving over time. 
Different kinds of graph evolve differently.
Some change very fast but some others not. 
For example, a telephone SS7 voice network generates server logs in each node every second, while an social interactive network modeled from customer reviews has no updates for days.
Zaki et el.\cite{Zaki2016} summarized that dynamic graph can be modeled as either \emph{Discrete Time Dynamic Graph (DTDG)} or \emph{Continuous Time Dynamic Graph (CTDG)} based on how the temporal information is expressed regarding to the evolution of dynamic graph. DTDG is a list of snapshots and each of them keeps the graph status at a certain moment. Meanwhile, CTDG can be viewed as a stream of graph updating event. 
The definitions of these two graph representation models will be given out in Sec.~\ref{sec-DTDG} and ~\ref{sec-CTDG} respectively.
Such modeling paradigms are well adopted in the community \cite{rossi2020temporal, Xu2019, xu2020inductive, kazemi2020representation}.

For a graph whose nodes or edges are frequently updated, it is more memory efficient to store it with DTDG modeling due to its snapshot-based representation\cite{masuda2020guide}.
However, it may induce some important temporal information loss if the observation frequency is not appropriately set.
For example, if strong periodicity presents in nodes' updating events, each node's activity reach the peak at noon everyday. But the observation frequency is set to be every 24 hours, then such periodicity pattern is never captured in the resulting snapshots. In comparison, CTDG can capture all temporal information as it is an event-based representation\cite{masuda2020guide,kempe2002connectivity}.

Due to their pros and cons, the most important decision to make is which storage model to use. Such decision must be made at the early stage of a dynamic graph learning use case. The selected storage model also limits the options of machine learning algorithms to only those are compatible. 
Algorithms that are designed specifically for DTDG, such as DySat\cite{sankar2020dysat} and STGCN\cite{yan2018spatial}, cannot be applied directly with CTDG storage model.
Similarly, algorithms such as TGAT\cite{xu2020inductive} that are designed for CTDG cannot be applied with DTDG either.
In order to offer a fast reference for practitioners the first hierarchy in our taxonomy to categorize different algorithms is the compatible graph storage model.

\subsection{Encoder And Decoder Learning Framework}
\begin{figure*}
\centering
\includegraphics[width=0.85\textwidth]{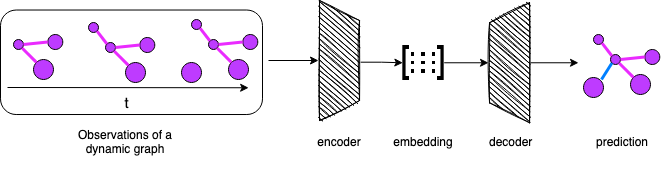}
\caption{Encoder-Decoder Learning Framework}
\label{ed-framework}
\end{figure*}

Graph learning algorithms are better considered to be under the encoder-decoder framework\cite{hamilton2017representation}. 
We follow the same framework when reviewing supervised dynamic graph learning algorithms.
As shown in Fig.~\ref{ed-framework}, the encoder turns the observations of a dynamic graph into its latent representation that is node-based.
This latent representation is called graph embedding.
The decoder decodes the generated graph embedding and give the prediction or classification result.
Under this framework, researchers can experiment on different encoder-decoder combinations to find the best one fitting the particular learning task \cite{kazemi2020representation}.

The modification of decoder to perform the three graph learning tasks is simple.
Node focus task is straightforward, since the embedding generated by encoder is node-based.
We can modify the decoder to take the concatenation of two nodes' embedding as input to perform edge focus task, which is referred as the pair-wise decoder\cite{hamilton2017representation}.
In the graph focus task, practitioners need to find an aggregation method to aggregate all nodes' embedding into the graph embedding. 
By modifying the decoder as described above, one can use the same encoder and decoder combination to perform all three dynamic graph learning problems. Therefore, algorithms fitting with such encoder-decoder framework are considered as general purposes learning methods.

\subsection{Implicit And Explicit Learning Model}
The last hierarchy of the taxonomy in this survey paper is whether a given algorithm uses time as an input feature explicitly or implicitly.
If time is used implicitly, it is not fed into the model as an independent feature. 
For example, the algorithm with a static graph encoder and LSTM decoder,  does not use time explicitly.
The success of temporal learning relies on the time based ordering of the input snapshots and the regular observation frequency.  
We refer this kind of learning model as \emph{Implicit Time Learning Model}.

Meanwhile, if a given algorithm takes time as an independent input feature, it is referred as \emph{Explicit Time Learning Model}.
Examples include TGAT\cite{xu2020inductive} and TGNN\cite{ma2020streaming} which use the time encoding Time2Vec\cite{kazemi2019time2vec} as a node feature.
Explicit Time Learning Model is capable of periodicity recognition and vector clock recognition, for which Implicit Time Learning Model is incapable.

\subsubsection{Periodicity Recognition}
In temporal data, periodicity is a frequent seen phenomena. 
For example, $50\%$ to $70\%$ of human movements can be explained by periodic behavior pattern\cite{cho2011friendship}.
Periodic pattern can be learnt by sequential model in DTDG setting.
The learning model infers the underlying timestamps by the ordering and position of the input snapshots. 
However, in CTDG, this is unfeasible without the explicit use of time.
Kazemi et al.\cite{kazemi2019time2vec} proposed Time2Vec to encode time and help the downstream learning model to recognize periodic pattern as well as linear time pattern.

TGAT\cite{xu2020inductive} applies time2vec in message passing to assign more weight to neighbors with similar periodic patterns to target nodes.
Explicit use of time with Time2Vec helps CTDG learning better learn periodical patterns.

\subsubsection{Vector Clock Recognition}
Information flowing between two given nodes needs to traverse their shortest path.
The difference in path length and edge property results in the difference in the flowing information's arrival time at the target node.
\textit{Temporal Distance} between starting nodes and destination nodes is one metric to evaluate how much time is needed for the information to flow from the starting one to the destination\cite{xuan2003computing}.
Because the temporal distance from the target node to its neighbors has different value, most up-to-date information with respect to its neighbors is sent at different timestamp.
This phenomena is described by the concept \textit{Vector Clock}\cite{lamport2019time}.
Since different nodes have different vector clocks,they evolve differently from each other.
\emph{Temporal Point Process(TPP)}\cite{rasmussen2011temporal} is one of the temporal learning methods that is capable to learn the impact from vector clock. 
With TPP and the explicit use of time, a learning algorithm could recognize the impacts from a given node's vector clock to its evolution.

Moreover, the explicit use of time makes time prediction tasks feasible in dynamic graph learning.
For the best of our knowledge, time prediction tasks could only be done with explicit time learning methods. Figure \ref{overview} shows the overall architecture of how this survey organizes different learning methods.

\comment{
Similar to correlation in statics, \emph{Temporal Correlation} is a measure to evaluate the importance of the hidden evolution model for its network evolution.
Namely, the degree of dependence between the next observation and the previous ones\cite{masuda2020guide}. 
There are multiple measures to evaluate the temporal correlation for both DTDG and CTDG.
\todo{add citations and temporal correlation for some frequent used dataset}.
By intuition, we believe the performance gain to use dynamic learning model vs. static graph learning model on dynamic graph is positively correlated to the given graph's temporal correlation, future work is to run experiment to verify this intuition and offer a fast tool for practitioners to decide which one to use in their learning task}

\section{Temporal Pattern Learning}
\label{time-model}
In this section, we present the \threestage.
This framework describes a general form of dynamic graph learning algorithms, how they learn and apply the temporal pattern for different graph learning tasks.
In some circumstances, node attributes are not available.
Such dynamic graphs are called \emph{non-attributed dynamic graph}. 
Conversely, dynamic graphs with attributes are called  \emph{attributed dynamic graph}. 
Previous works are limited to either attributed or non-attributed dynamic graphs\cite{toivonen2009comparative, farine2017dynamics, artime2017dynamics }. 
For the best of our knowledge, \threestage~is the only framework which could be generalized to both attributed and non-attributed dynamic graph learning.
Subsection \ref{three-stages} presents the idea of the proposed framework and briefly introduces its three stages.
Subsection \ref{attribute-update}, \ref{association-process} and \ref{mp} explain the three stages in detail.
Subsection \ref{generalization} explains how to apply \threestage~ in attributed and non-attributed dynamic graphs.
The equations presented in this section assumes the given learning task is an extrapolation task.
With some modification, they can be applied to interpolation tasks as well.

\subsection{\threestage}
\label{three-stages}
\emph{\threestage} describes how a learning algorithms learns the temporal pattern.
A temporal pattern is a repetitive pattern in the given dynamic graph's evolution.
A particular learning algorithm could learn and use it to perform those mentioned graph learning tasks.
Given that the state of a dynamic graph $G_t$ at time $t$ is described by a timestamped state tuple $(E_t, \matr{X_t}, t)$, in which $E_t$ and $\matr{X_t}$ is the edge connections and node attributes matrix observed at time $t$. 
The temporal pattern could be expressed the following function:
\begin{equation}
(\hat{E}_{t_{n+1}}, \matr{\hat{X}}_{t_{n+1}}, t_{n+1}) = \text{tp}(\{E_{T}\}, \{\matr{X_{T}}\}, T)
\label{e-pattern}
\end{equation}
where $\{E_{T}\}$ and $\{\matr{X_{T}}\}$ are the set of observations of $E$ and $\matr{X}$ in time period $T = [t_0,t_1,\cdots,t_n]$. 
$\hat{E}_{t_{n+1}}$ and $\matr{\hat{X}_{t_{n+1}}}$ are the prediction of $E$ and $\matr{X}$ for timestamp $t_{n+1}$ based on the observed history.

To predict any missing entry in the state tuple, a learning algorithm needs an output function $\text{out}(\cdot)$ which takes the predicted state $(\hat{E}_{t_{n+1}}, \matr{\hat{X}_{t_{n+1}}}, t_{n+1})$ and the known state $(E_{t_{n+1}}-e_{\{i,j\},t_{n+1}}),\matr{X_{t_{n+1}}},t_{n+1})$ as input.
For example, in edge focus task, to predict the status of a given edge  $e_{\{i,j\},t_{n+1}} \in E_{t_{n+1}}$ between $v_i$ and $v_j$,  
the output function $\text{out}(\cdot)$ could be written as:
\begin{equation}
\begin{split}
  e_{\{i,j\},t_{n+1}} &= \text{out}((\hat{E}_{t_{n+1}}, \matr{\hat{X}}_{t_{n+1}}, t_{n+1}),\\ &(E_{t_{n+1}}-e_{\{i,j\},t_{n+1}}),\matr{X_{t_{n+1}}},t_{n+1}))
\label{output-function}
\end{split}
\end{equation}
A supervised dynamic graph learning algorithm needs to learn the temporal pattern $\text{tp}(\cdot)$ from the ground true history.

\threestage~ assumes $\text{tp}(\cdot)$ to be a three stages process such that it is a composite of the three functions:
\begin{subequations}
\label{three-stage-func}
    \begin{align}
    \label{self-update-func}
   \matr{X'}_{t_{n+1}} &= \text{asu}(\{\matr{X}_{T}\},T, t_{n+1})
   \\
   \label{associate-func}
   \matr{\hat{E}}_{t_{n+1}} &= \text{ap}(\{E_{T}\},\{\matr{X}_{T}\},\matr{X'}_{t_{n+1}}, T, t_{n+1})\\
  \label{message-passing} 
   \matr{\hat{X}}_{t_{n+1}} &= \text{mp}(\hat{E}_{t_{n+1}},\matr{X'}_{t_{n+1}}) \\
   \label{composite}
   \text{tp} &= \text{mp} \circ \text{ap} \circ \text{asu}
    \end{align}
\end{subequations}
Functions $\text{asu}(\cdot)$, $\text{ap}(\cdot)$ and $\text{mp}(\cdot)$ each represents an intermediate stage in a dynamic graph's evolution which will be detailed in following subsections.
The operation $\circ$ is function composition.

\begin{figure}[t] 
\centering
\includegraphics[width=0.47\textwidth]{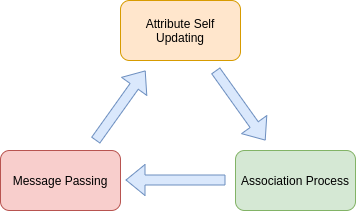}
\caption{Three Stages Recurrent Temporal Learning Model}
\label{temporal_process}
\end{figure}

As shown in Fig. \ref{temporal_process}, the first stage is the \emph{Attribute Self-Updating} defined as in Eqn.~(\ref{self-update-func}). 
This stage captures the impact from the external factors to the graph evolution and gives out $\matr{X'_{t_{n+1}}}$ as an estimation of $X$ at $t_{n+1}$.
Here we present a model with only node attributes self-updating.
Readers can extend it to edge attributes and graph attributes self-updating with similar schema.
The change of attributes triggers the second stage named \emph{Association Process}.
As in Eqn.~(\ref{associate-func}), the association process describes the evolution of connection pattern.
It generates new connection pattern based on the current connection pattern, timestamp and the self-updated attributes.
The new connection pattern triggers the last stage which is the  \emph{Message Passing}.
As defined as Eqn.~(\ref{message-passing}), this stage integrates the impact of attributes self-updating and association process to generate the attributes for the next state. 
Consequently, the result of message passing would be the starting point of attributes self-updating in the next round of the evolution. 
A dynamic graph's evolution can be described as a recurrent process of these three stages.   

\subsection{Attributes Self-Updating}
\label{attribute-update}
The first stage in \threestage is the attributes self-updating.
This stage captures the impact of external factors to the evolution of the given dynamic graph's attribute.
Depending on the context, the attributes being impacted could be node attributes, edge attributes or graph attributes.
The change of attributes alternatively drives the evolution of the given dynamic graph's connection pattern\cite{toivonen2009comparative}.
As an example, in a client-item knowledge graph, the unobserved change of the client's status would triggers the change of his interested items.

\begin{defn} \textit{
Attributes Self Updating: the change of node, edge or graph's attributes resulted from external factors. Its equation $\text{asu}(\cdot)$ is defined in Eqn.~(\ref{self-update-func})}   
\end{defn}

As in our example, this stage only takes input the history of the node attributes and the timestamps.
Its output $\matr{X'}_{t_{n+1}}$ is the estimated node attributes for the next timestamp $t_{n+1}$. 

\subsection{Association Process}
\label{association-process}
The change of attributes triggers the development of a new connection pattern. 
The process that describes the evolution of a dynamic graph's connection is called Association Process.

\begin{defn} \textit{
Association Process:  The process that a particular dynamic graph develops, abandons or modifies the edges between its nodes. Its equation $\text{ap}(\cdot)$ is defined in Eqn.~(\ref{associate-func}).}
\end{defn}

As shown in Eqn.~(\ref{associate-func}), the association process outputs the future connection pattern based on the given dynamic graph's evolution history, the estimated future attributes from the first stage, and the timestamps.

\subsection{Message Passing}
\label{mp}
In graph analysis, we believe nodes are impacted by their neighbours.
To learn how nodes are impacted by their neighbours, \emph{Message Passing} is developed for Graph Neural Network(GNN)s\cite{gilmer2017neural, hamilton2017inductive,velivckovic2017graph}.

\begin{defn} \textit{
Message Passing\cite{gilmer2017neural},  also known as Affinity Propagation\cite{wang2008adaptive, dueck2009affinity} and Communication\cite{trivedi2019dyrep}, is a local neighborhood information aggregation method, which updates node attributes by aggregating messages received from neighboring nodes and the connected edges. Its equation $\text{mp}(\cdot)$ is defined in Eqn.~(\ref{message-passing})}
\end{defn}

Recent advancements in DTDG learning apply message passing to generate node-based graph embedding.
The resulted graph embedding is considered to be a latent representation of the underlying graph's network structure and nodes/edges attributes\cite{hamilton2017representation, wu2020comprehensive}.
Therefore, it contains important information for graph related machine learning tasks.

\subsection{Generalization To Attributed And Non-attributed Dynamic Graphs}
\label{generalization}
\threestage~ describes how attributed dynamic graph evolves by Eqn.~(\ref{three-stage-func}).
Supervised dynamic graph learning algorithms learns the temporal pattern $F$ by optimizing the trainable weights in Eqn.~(\ref{three-stage-func}) and (\ref{output-function}) to best fit the input history.

When generalized to non-attributed dynamic graph, the impact from node attributes are usually ignored since they are not available.
The only driver considered in the graph's evolution is the association process.\cite{masuda2020guide,toivonen2009comparative}.
In this case we can drop Eqn.~(\ref{self-update-func}) from Eqn.~(\ref{composite}) and setting Eqn.~(\ref{associate-func}) to take input only $\{E_T\}$ and the timestamps as follows:

\begin{equation}
\hat{E}_{t_{n+1}} = \text{ap}(\{E_{T}\}, T, t_{n+1})
\label{non-attributed-model}.
\end{equation}

TGN\cite{rossi2020temporal} and APAN\cite{wang2021apan} adopts \threestage~ by applying the node memory units to capture the attribute self-updating.

Know-Evolve\cite{trivedi2017know}, DyRep\cite{trivedi2019dyrep} and TGAT\cite{xu2020inductive} adopt \threestage~ by setting Eqn.~(\ref{self-update-func}) to always return the last observation in the input attributes history $\{\matr{X}_T\}$.

The learning of temporal pattern $F$ is achieved by applying temporal learning algorithms in the dynamic graph encoder or  directly in the decoder.
Commonly-used temporal learning algorithms include RNN family neural networks, 1-D convolutional networks, time series analysis methods and attention networks.
Recent trend is to explore the use of Temporal Point Process in temporal learning\cite{trivedi2017know, trivedi2019dyrep}.

\section{Discrete Time Dynamic Graph Learning}
\label{sec-DTDG}
Dynamic network represented by DTDG has a discretized time dimension.
Each observation in the given dynamic graph $G_T$ is expressed as a snapshot of the given graph attached with the observation timestamp.
DTDG is defined as follow:

\begin{defn} 
\emph{Discrete Time Dynamic Graph (DTDG)}: \textit{a dynamic graph $G_{T} = O_{T}$ for a time span $T=[t_1:t_n] $is stored as DTDG, if each stored observation $o_{t_i}$ in $O_{T}$ is a snapshot of the given graph $o_{t_i} = (V_{t_i}, E_{t_i}, \matr{X}_{t_i})$ where $V_{t_i}$, $E_{t_i}$ and $\matr{X}_{t_i}$ are nodes, edges and node features matrix observed at $t_i$ }.
\end{defn}

The data pipeline that makes the observation and stores the snapshot in a database is usually running in a regular frequency depending on the requirement, such as once per hour or once per day.
The timestamps are usually as simple as ordered integers instead of actual date and time (e.g., $T = [1,2,3, \dots, n]$).

\begin{figure*}[ht]
\begin{subfigure}{.33\textwidth}
  \centering
  \includegraphics[width=0.65\textwidth]{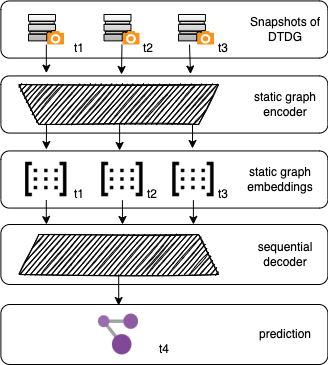}
\caption{Static Graph encoder - Sequential Decoder Framework For DTDG}\label{DTDG-framework}
\end{subfigure}
\begin{subfigure}{.33\textwidth}
\centering
\includegraphics[width=0.65\textwidth]{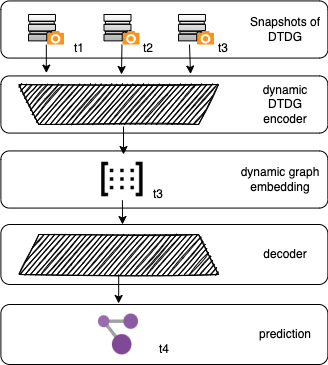}
\caption{Dynamic Graph Encoder - Simple Decoder Framework For DTDG}\label{DTDG-dynamic}
\end{subfigure}
\begin{subfigure}{.33\textwidth}
\centering
\includegraphics[width=0.65\textwidth]{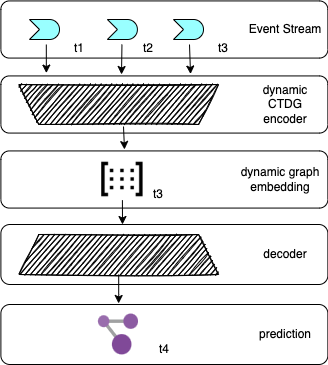}
\caption{Dynamic Graph Encoder - Simple Decoder Framework For CTDG}
\label{ctdg-learning}
\end{subfigure}

\caption{Different Encoder-Decoder Architectures For Dynamic Graph}
\end{figure*}

The most seen form of supervised DTDG learning is the static graph encoder - sequential decoder framework.
As shown in Fig.~\ref{DTDG-framework}, Algorithms following this framework use a static graph encoder to generate embedding for each snapshot and pass those embedding to a supervised sequential decoder for inference.
Because the dynamic network has already been sliced into snapshots, there is no explicit usage of temporal information in the encoder.
The encoder only captures the graph structure, property and attributes for each snapshot. 
Temporal information is learnt through the sequential decoder.

There are some emerging attempts to learn temporal pattern as well as graph topology and attributes in the encoder. 
These algorithms follow the dynamic graph encoder - simple decoder framework as shown in Fig.~\ref{DTDG-dynamic}.
Instead of generating node-wise graph embedding for each snapshot in each inference, the dynamic graph encoder only generates one embedding recursively based on inputs in the past and the current input snapshot at each inference.
Table \ref{tab:encoders-dtdg} lists all encoders for supervised DTDG learning  in this survey.
So far as we summarize, all supervised DTDG learning methods do not learn the graph attribute self updating process. 

	\begin{table*}
		\centering
		\caption{DTDG Graph Encoders}
		\label{tab:encoders-dtdg}
		\begin{tabular}{c|l|l} 
		\hline
			Time Model & Graph Type & Encoder \\
			\hline
		  Implicit & non-attributed static graph & Shallow Embedding\cite{belkin2003laplacian,ahmed2013distributed,cao2015grarep,ou2016asymmetric,wang2017community,perozzi2014deepwalk,yang2015network,chen2018harp,grover2016node2vec,tang2015line}\\\cline{3-3}
			
			& & autoencoder Based Embedding\cite{cao2016deep, wang2016structural} \\
			  \cline{2-3}
			&attributed static graph &Message Passing GNN\cite{scarselli2008graph,Gilmer2017,Zhou2018,hamilton2017inductive}\\\cline{3-3}
			&& Graph Convolution Network\cite{kipf2016semi,kipf2016variational,zheng2019addgraph,hamilton2017inductive,niepert2016learning,manessi2020dynamic,mohanty2018graph}\\\cline{3-3}
             
            && Graph Attention Network\cite{velivckovic2017graph,xu2015empirical,sankar2020dysat} \\
			\cline{2-3}
			&dynamic graph & Kalman Filter Based Encoder\cite{sarkar2007latent} \\ \cline{3-3}
			&& EvolveGCN\cite{pareja2020evolvegcn}\\ \cline{3-3}
			&& Spatial-Temporal Graph Convolution Network\cite{yan2018spatial}\\\hline
			Explicit & N.A. & N.A.\\\hline
		\end{tabular}
	\end{table*}

\subsection{Static Graph Encoder}
\subsubsection{Non-Attributed Static Graph Encoder}
Static graph encoders could be categorized as non-attributed and attributed graph encoder. 
Non-attributed static graph encoders are not widely discussed in academia these days, due to their incapability to leverage graph attributes.
This motivates the recent exploration in attributed static graph embedding approaches.
We will first review briefly the non-attributed static graph embedding methods, and then provide an in-depth review of attributed static graph embedding methods.   
For those interested in a more detailed non-attributed static graph representation learning algorithms, we refer them to read the recent work of Hamilton et al. \cite{hamilton2017representation}, Cui et al.\cite{cui2018survey} and kazemi et al.\cite{kazemi2020representation}.

A basic form of non-attributed static graph encoder is the shallow embedding approach, which aims to transform graph structure and property to node-level graph embedding \cite{cui2018survey}.
However, graph attributes are not considered in shallow embedding.
Shallow embedding includes
\begin{itemize}
    \item Matrix Factorization based approaches such as \emph{Laplacian Eigenmaps}\cite{belkin2003laplacian}, \emph{Graph Factorization}\cite{ahmed2013distributed}, \emph{GraRep}\cite{cao2015grarep}, \emph{HOPE}\cite{ou2016asymmetric} and M-NMF\cite{wang2017community};
    \item Random walk based approaches such as \emph{DeepWalk}\cite{perozzi2014deepwalk}, \emph{TADW}\cite{yang2015network},  \emph{HARP}\cite{chen2018harp} and \emph{node2vec}\cite{grover2016node2vec}.
    \item Other approaches, such as \emph{LINE}\cite{tang2015line} .
\end{itemize}

Shallow embedding encoder is simply an embedding look up based on node ids, there is no parameter sharing between nodes.
Hence, the computation is inefficient and the trained encoder cannot be used for new graphs with unseen nodes \cite{hamilton2017representation}.

To overcome these challenges and leverage graph attributes in the embedding generating process, multiple approaches are proposed to parameterized graph embedding (i.e, parameter sharing between nodes).

\emph{Deep Neural Graph Representation (DNGR)}\cite{cao2016deep} and \emph{Structural Deep Network Embeddings (SDNE)}\cite{wang2016structural} apply \emph{autoencoder}\cite{rumelhart1986learning} to map a high-dimension node similarity matrix to a low-dimension node embedding. These two approaches enable sharing of parameters, which can provide an efficient computation and be applied with unseen nodes. 
\subsubsection{Attributed Static Graph Encoder}
To leverage graph attributes in static graph embedding generation, attributed static graph encoders are proposed.
Attributed static graph encoders are based on \emph{Graph Neural Network (GNN)}\cite{scarselli2008graph}. 
Such algorithms follow a message passing schema to aggregate neighborhood information and generate embedding for the target node \cite{Gilmer2017}.
In this survey, we focus on some widely used GNN based attributed graph encoders which are the foundation of dynamic graph encoders to review in later section.

GNN assigns state $\vecr{h}_{v_i}$ to node $v_i$ of the input graph.
Given mappings $N(v_i)$ to be all neighbors of node $v_i$,
mapping $E(v_i)$ to be all edges connecting node $v_i$ and its neighbors $N(v_i)$, $f_w(\cdot)$ to be the neighborhood information aggregating function, 
state $\vecr{h}_{v_i}$ at the $k$ layer is defined as in Eqn.~(\ref{GNN-state}). 
\begin{equation}
    \label{GNN-state}
   \vecr{h}_{v_i}^k = f_w(\vecr{h}^{k-1}_{v_i}, E(v_i),\vecr{h}^{k-1}_{N(v_i)}),
\end{equation}
which can be written as Eqn.~(\ref{non_positional}) when there is no positional information for the neighbours. 
\begin{equation}
\label{non_positional}
\vecr{h}_{v_i}^k = \sum_{u\in N(v_i)}f_w(\vecr{h}^{k-1}_{v_i}, e(v_i, u),\vecr{h}^{k-1}_{u}).
\end{equation}

The output state from the last layer $n$ is the graph embedding $\vecr{z}_{v_i}$:
\begin{equation}
\label{final-embedding}
\vecr{z}_{v_i} = \vecr{h}_{v_i}^n. 
\end{equation}

This state updating process is described as a message passing process in \cite{Gilmer2017,Zhou2018}.
In each iteration, message from each node is passed through the edges to their neighbours.
To learn local neighborhood structure, $k$ is required to be a small value, such as $k=2$ in the learning of $2$-hop neighborhood information.
With Banach’s fixed point theorem, the state values will converge with the update of state in iterations~\cite{banach1922operations, Gilmer2017}. 
Therefore, to learn the global graph structure, we can just continue the iterative updating until the change in the state value between two consecutive iterations is close to 0 as shown in Eqn.~(\ref{converge-condition}).
When the state values converge, the global graph structure and property is embedded in the resulting embedding.

\begin{equation}
\label{converge-condition}
\vecr{h}_{v_i}^k - \vecr{h}_{v_i}^{k-1} \approx 0.
\end{equation}

Stacking the converged state value  $z$ of each node together to produce $Z$, we obtain the node-wise embedding of the input graph. 
\emph{GraphSAGE}\cite{hamilton2017inductive} and \emph{column network}\cite{pham2017column} are following the same schema with different neighborhood information aggregation functions.

Inspired by GNN,  \emph{Graph Convolution Network (GCN)}\cite{kipf2016semi} is a generalization of \emph{Convolution Neural Network (CNN)}\cite{krizhevsky2012imagenet} to static graph data structure with spectral method.
GCN applies graph Laplacian to generate the feature maps which are shared within the whole graph as in CNN.
Given the adjacency matrix $\matr{A}$, $\matr{\hat{A}} = \matr{A}+ \matr{I}_N$, the diagonal degree matrix $\matr{\hat{D}} : \matr{\hat{D}}_{ii}=\sum_{j}\matr{\hat{A}}_{ij}$, the graph Laplacian is calculated as follows:
\begin{equation}
\label{laplacian}
    \matr{L} = \matr{\hat{D}}^{\frac{1}{2}}\matr{\hat{A}}\matr{\hat{D}}^{\frac{1}{2}}.
\end{equation}
Moreover, a convolution layer with the output of $d$ feature maps is calculated as follows:
\begin{equation}
\label{gcn}
    \matr{Z} = \matr{L}\matr{X}\matr{W},
\end{equation}
where $\matr{X} \in \mathcal{R}^{|V| \times d}$ is the input node feature matrix with $d$ features and $\matr{W} \in \mathcal{R}^{d \times d'}$ is the layer weight matrix to generate $d'$ feature maps.
The output $\matr{Z}$ of a hidden convolution layer is usually passed to an $\text{ReLU}$ activation function, and the output of activation function will then be passed to the next layer as the input. 
As an example, a classic two layers GCN with $\text{softmax}$ activation in the output layer has the form in Eqn.~(\ref{2layer-gcn}).
\begin{equation}
\label{2layer-gcn}
    \matr{Z} = \text{softmax}(\matr{L} \cdot \text{ReLU}(\matr{L}\matr{X}\matr{W}^{\text{hidden}}) \cdot \matr{W}^{\text{out}}).
\end{equation}

One example of GCN as static graph encoder to learn the neighborhood structure for each snapshot and a sequential decoder to learn the temporal pattern between snapshots is AddGraph\cite{zheng2019addgraph} .

Similar to GCN, \emph{Graph Attention Network (GAT)} \cite{velivckovic2017graph} applies the attention mechanism \cite{bahdanau2014neural,vaswani2017attention} to determine the importance of neighboring nodes in the neighborhood aggregation for the target node:
\begin{equation}
\label{general-attention}
    \text{attn}(\matr{Q}, \matr{K}, \matr{V}) = \text{softmax}(\text{score}(\matr{Q},\matr{K}))\matr{V},
\end{equation}
where $\vecr{Q}$ is the linear projection of the target node's input state $\vecr{h}_{v_i}$ from the previous layer. $\matr{K}$ and $\matr{V}$ are the linear projections of the input state of the $v_i$'s neighboring nodes:
\begin{equation}
\begin{split}
\vecr{Q} &= \vecr{h}_{v_i} \vecr{W}_q\\
\matr{K} &= [\vecr{h}_{v_1},\vecr{h}_{v_2},\cdots,\vecr{h}_{v_j},\cdots,\vecr{h}_{v_n}]\matr{W}_k,  \forall v_j\in N(v_i)\\     
\matr{V} &= [\vecr{h}_{v_1},\vecr{h}_{v_2},\cdots,\vecr{h}_{v_j},\cdots,\vecr{h}_{v_n}]\matr{W}_v,  \forall v_j\in N(v_i) 
\end{split}
\end{equation}

The function $\text{score}(\cdot)$ calculates a score representing how well $\vecr{Q}$ align to $\matr{K}$. 
The alignment scores for $v_i$'s neighborhood $N(v_i)$ are then normalized by a $\text{softmax}$ layer. 
The output of the whole attention layer is the product of the normalized alignment score and the linear projection $\matr{V}$.

In \emph{Transformer} \cite{vaswani2017attention}, the score function is the dot product of $Q$ and $K$,  while  the score function is a leaky ReLU layer in  \emph{Graph Attention Network (GAT)}\cite{velivckovic2017graph,xu2015empirical}.
GAT is used as static graph encoder to learn DTDG in Dysat\cite{sankar2020dysat}.
All attributed static graph encoders can be used to generate embedding in a supervised or unsupervised manner\cite{hamilton2017inductive,kipf2016variational,kipf2016semi,velivckovic2017graph, pham2017column}.

\subsection{Sequential Decoder}
In supervised DTDG learning, the temporal pattern is revealed by the change between snapshots along the time dimension. 
By sorting the static graph embedding generated from the series of  snapshots according to their timestamps and treating them as sequential data, temporal pattern can be learnt via a sequential decoder.
However, sequential decoder does not use time explicitly and is not capable to perform time predicting tasks.
Table \ref{tab:decoders} lists different kind of  decoders summarized in this work.
	\begin{table*}
		\centering
		\caption{Decoders}
		\label{tab:decoders}
		\begin{tabular}{c|l|l} 
		\hline
			Storage Model & Time Model & Decoder \\
			\hline
			DTDG/CTDG & Implicit & Time Series Analysis Methods \cite{da2012time,huang2009time,gunecs2016link}\\ \cline{3-3}
			&&RNN family\cite{rumelhart1986learning,hochreiter1997long,seo2018structured,niepert2016learning,manessi2020dynamic,mohanty2018graph,yuan2017temporal,taheri2019learning,bogaerts2020graph} \\\cline{3-3}
			&&1-D Convolution Network \cite{wang2021graphtcn}\\\cline{3-3}
			&&Positional Temporal Self Attention\cite{sankar2020dysat}\\\hline
			CTDG &  Explicit & Temporal Point Process Decoder\cite{trivedi2017know,chang2020continuous}\\\hline
					\end{tabular}
	\end{table*}
\subsubsection{Time Series Analysis Methods}
DTDG can be considered as a time series of snapshots. 
Hence, traditional time series methods can be used naturally as decoders to learn temporal patterns. 
\emph{Exponential Moving Average (EMA)} and \emph{Auto-Regressive Integrated Moving Average (ARIMA)} are two widely-used methods, which are used as decoders in previous works
\cite{da2012time,huang2009time,gunecs2016link}.
As an example, Eqn.~(\ref{ema}) shows the equation of EMA with a predefined smoothing factor $\alpha \in (0,1)$.

\begin{equation}
\centering
    \matr{Z}_{t_{n+1}} = \sum_{i=0}^{n}\alpha(1-\alpha)^{i}\matr{Z}_{t_{n-i}}.
    \label{ema}
\end{equation}

\subsubsection{Recurrent Neural Network (RNN) family}
RNN based methods are frequently-used supervised sequential decoder and are capable to learn temporal correlation.
A basic form of RNN decoders is defined by a recurrent state function which takes the previous state and the current embedding as input\cite{rumelhart1986learning}:
\begin{equation}
    \label{rnn-equation-h}
    \vecr{h}_{v_i,t_n} = \text{RNN}(\vecr{h}_{v_i,t_{n-1}}, \vecr{z}_{v_i,t_{n}}).
\end{equation}
The generated state is used as the input in the output unit which is usually a \emph{Feed forward Neural Network (FNN)}:
\begin{equation}
    \label{rnn-equation-o}
    o_{v_i,t_n} = \text{FNN}(\vecr{h}_{v_i,t_n}).
\end{equation}

One frequently used RNN model is the \emph{Long Short Term Memory (LSTM)}\cite{hochreiter1997long} which better learns long-term temporal pattern. 
For example, Seo et al.\cite{seo2018structured} applied the spectral GCN\cite{defferrard2016convolutional} as static graph encoder and LSTM as decoder;
Narayan et al.\cite{narayan2018learning} used a different GCN\cite{niepert2016learning} as static graph encoder and LSTM as decoder; Manessi et al.\cite{manessi2020dynamic} and Mohanty et al.\cite{mohanty2018graph} also used different version of GCN as static graph encoder and LSTM as decoder to perform dynamic graph learning. 
Yuan et al.\cite{yuan2017temporal} applies message passing GNN as static graph encoder and a four gates LSTM as sequential decoder to learn the dynamic graph constructed from video frames and performs object detection. 
DyGGNN\cite{taheri2019learning} uses a gated graph neural network\cite{li2015gated} as encoder and LSTM as decoder.
Bogaets et al.\cite{bogaerts2020graph} applies CNN as static graph encoder and LSTM as sequential decoder in traffic forecasting for road network in the city of Xi'an.

\subsubsection{1-D Convolution Neural Network (Conv1d)} Conv1d is used in time series analysis\cite{lecun1995convolutional}.
Therefore, it is naturally to use as the sequential decoder to learn the temporal pattern.
Unlike RNN families, Conv1d only learns short-term temporal patterns in the given time frame. 
Periodicity of those short term temporal patterns could be learnt by stacking multiple Conv1d layers.
By carefully setting the size of each feature map, Conv1d inputs the embedding of a given node for $n$ most recent snapshots $[\vecr{z}_{v_i,t_1},\vecr{z}_{v_i,t_2},\cdots, \vecr{z}_{v_i,t_n}]$, and generates the decoded state $\vecr{h}_{v_i,t_{n}}$ for node $v_i$ at time $t_n$:

\begin{equation}
    \label{cnn-equation-h}
    \vecr{h}_{v_i,t_n} = \text{Conv1d}([\vecr{z}_{v_i,t_1},\vecr{z}_{v_i,t_2},\cdots, \vecr{z}_{v_i,t_n}]).
\end{equation}
The decoded state is then pass to the output unit as in Eqn.~(\ref{rnn-equation-o}).

GraphTCN\cite{wang2021graphtcn} applies a GAT based static graph encoder and Conv1d as the sequential decoder to learn the spatial and temporal information in Human Trajectory Prediction.

\subsubsection{Temporal Self-Attention (TSA)} 
Attention mechanism is proved to perform very well in sequential data learning \cite{bahdanau2014neural, chen2021continuous}.
As in Eqn.~(\ref{tsa-equation-q}), given the node embedding $\matr{Z}_{v_i}$ of the target node $v_i$, each element encodes a snapshot from the observation period $[t_1, t_2, \cdots, t_{n-1}, t_n]$, TSA uses each element in $\matr{Z}_{v_i}$ as a query to attend over the whole input history $\matr{Z}_{v_i,[t_1:t_{n}]}$ to generate the temporal graph embedding for the corresponding snapshot. The output of a TSA layer $\matr{H}_{v_i}$ has the same time dimension as the input, such that it is feasible to stack multiple TSA layers to learn the evolution of its local neighborhood structure and attributes over time.
\begin{equation}
    \begin{split}
         \matr{Z}_{v_i} &= [\vecr{z}_{v_i,t_1},\vecr{z}_{v_i,t_2},\cdots, \vecr{z}_{v_i,t_n}]\\
         \vecr{Q}_{t_i} &= (\vecr{z}_{v_i,t_i} + \matr{P}_{t_i}) \matr{W}_q,\\
         \matr{K} &= (\matr{Z}_{v_i}+\matr{P}) \matr{W}_k,\\
         \matr{V} &= (\matr{Z}_{v_i}+\matr{P}) \matr{W}_v,\\
         \vecr{h}_{v_i,t_i} &= \text{attn}(\vecr{Q}_{t_i}, \matr{K},\matr{V}),\\
         \matr{H}_{v_i} &= [\vecr{h}_{v_i,t_1}, \vecr{h}_{v_i,t_2}, \cdots,\vecr{h}_{v_i,t_n}].
    \end{split}
    \label{tsa-equation-q}
\end{equation}

As in Eqn.~(\ref{general-attention}), Dysat\cite{sankar2020dysat} applied TSA as the sequential decoder in DTDG learning.
Their score function is shown in Eqn.~(\ref{tsa-score}). 

\begin{equation}
    \label{tsa-score}
    \text{score}(\vecr{Q}_{t_i},\matr{K}) = \frac{\vecr{Q}_{t_i}\matr{K}^{\textbf{T}}}{\sqrt{d'}}. 
\end{equation}

To the best of our knowledge, there is no explicit time sequential decoder proposed for supervised DTDG learning.
Because the snapshots are taken in a regular frequency, temporal information is revealed in the ordering and the position of the snapshots.
However, without using time as a learning feature, implicit time sequential decoder is not capable to perform time prediction tasks as discussed in Sec.~\ref{ML-dg}.

\subsection{Dynamic DTDG Encoder}

\subsubsection{Implicit Time DTDG Encoder}
Recent developments commonly use attributed dynamic graph embedding as an encoder to learn both graph structures and temporal correlations.
The generated dynamic graph embedding is then fed to a simple supervised predictive model as an decoder to conduct prediction.

\paragraph{Kalman Filter Based Encoder} 
\emph{Kalman Filter}\cite{kalman1960new}, also called \emph{Linear Quadratic Estimation}, is widely used in sensor signal refinement.
It is efficient in handling uncertainty caused by random external factors.
When considering node properties as sensor measures, Kalman Filter can be used to generate dynamic node embedding.
A hidden state matrix $\matr{H}_{t-1}$ at time $t-1$ is formed by stacking the hidden state of each node in the graph or the local neighborhoods.
Kalman Filter based encoder is a two-step recurrent process that includes the prediction step and the hidden state updating step. 
Given the hidden state $\matr{H}_{t-1}$ at snapshot $t-1$ and its estimated covariance matrix $\matr{\hat{P}}_{t-1}$, the embedding matrix $\matr{Z}_t$ at time $t$ and its predicted covariance matrix $\matr{P}_t$ is calculated as in Eqn.~(\ref{kalman-prediction-state}), where $\matr{W}$ and $\matr{B}$ are trainable parameters, and $\matr{Q}_t$ is a random noise drawn from a zero-mean Gaussian distribution, and $N_{t-1}$ is a control factor that could be simply 0 as in Sarkar et al.\cite{sarkar2007latent} or neighborhood embedding aggregation.

\begin{equation}
\begin{aligned}
  \matr{Z}_t &= \matr{WH}_{t-1} + \matr{BN}_{t-1}, \\
  \matr{P}_t &=\matr{W\hat{P}}_{t-1}\matr{W}^{\textbf{T}} + \matr{Q}_t.
\end{aligned}
\label{kalman-prediction-state}
\end{equation}

Once a new observation of node attributes $\matr{X}_t$ at time $t$ is obtained, the Kalman gain $\matr{K}_t$ is defined as in Eqn.~ (\ref{kalman-gain}), where $\matr{J}$ is a trainable parameter. 

\begin{equation}
    \label{kalman-gain}
    \matr{K}_t = \matr{P}_t\matr{J}^{\textbf{T}}(\matr{JP}_t\matr{J}^{\textbf{T}} + \text{Cov}(\matr{X}_t))^{-1}.
\end{equation}

The hidden state $\matr{H}_t$ and the estimated covariance $\matr{\hat{P}}_t$ are updated as follows:

\begin{equation}
\begin{aligned}
  \matr{H}_t &= \matr{Z}_t + \matr{K}_t(\matr{X}_t - \matr{JZ}_t), \\
  \matr{\hat{P}}_t &= \matr{P}_t - \matr{K}_t \matr{J P}_t.
\end{aligned}
\label{kalman-update-state}
\end{equation}

\paragraph{EvolveGCN}
The idea of EvolveGCN\cite{pareja2020evolvegcn} is simple and interesting.
In order to to make the model adaptable to new added nodes, EvolveGCN focuses on training an RNN model to learn the temporal dynamic presented in the underling GCN. 
Namely, the parameters in the underling GCN are not learned. 
They are predicted by an RNN model.
There are two versions of the EvolveGCN unit, which are the hidden unit \emph{EGCU-H} and the output unit \emph{EGCU-O}. 
\emph{EGCU-H} takes the input of last layers output states $\matr{H}^{l-1}_t$ and the parameters from the last time step $\matr{W}^l_{t-1}$, and then outputs the parameter for the current time step $\matr{W}^l_t$ as shown in Eqn.~(\ref{evolvgcn-h}).

\begin{equation}
    \label{evolvgcn-h}
    \begin{split}
        \matr{W}^l_t &= \text{GRU}(\matr{H}^{l-1}_t, \matr{W}^l_{t-1}),\\
        \matr{H}^{l+1}_t &= \text{GNN}(\matr{A}_t,\matr{H}^{l}_t, \matr{W}^l_{t}).
    \end{split}
\end{equation}

Similarly, \emph{EGCU-O} calculates the parameter for the underling GCN at the current time step by LSTM, but takes only the parameter from the last time step as input as follows:

\begin{equation}
    \label{evolvgcn-o}
    \begin{split}
        \matr{W}^l_t &= \text{LSTM}(\matr{W}^l_{t-1}),\\
        \matr{H}^{l+1}_t &= \text{GNN}(\matr{A}_t,\matr{H}^{l}_t, \matr{W}^l_{t}).
    \end{split}
\end{equation}

\paragraph{Spatial Temporal Graph Convolutional Network (ST-GCN)}
ST-GCN is developed to learn both the spatial and temporal patterns for human action recognition\cite{yan2018spatial}. 
An ST-GCN layer is composed by two operations: the spatial convolution and temporal convolution. 
Spatial convolution learns the graph structure pattern.
It adds a partitioning strategy to spectral GCN as described in Eqn.~\ref{gcn} to assign different weights to the nodes in different partitions, so as to learn the feature importance for different partitions based on their spatial information. 
There are three partitioning strategies proposed in \cite{yan2018spatial}: 

\begin{itemize}
    \item \emph{uni-labeling}: all nodes are assigned in the same partition. 
    \item \emph{distance partitioning}: the target node $v_i$ is assigned to partition $0$, and the partitioning of its neighbors is determined by the length of their shortest paths to $v_i$.
    \item \emph{spatial configuration}: the partition is determined by the given node's distance to the graph's centroid, as shown in Eqn.~(\ref{spatial-partition}), where $r_i$ is the distance between $v_i$ and the graph centroid, and $l_{ti}(\cdot)$ is a function whose output is the partition label of the input node $v_j$ at time $t$ in the state calculation for node $v_i$.
\end{itemize}

\begin{equation}
    \label{spatial-partition}
    l_{ti}(v_{tj}) = \begin{cases}
     0 & \text{if $r_j = r_i$}\\
     1 & \text{if $r_j < r_i$}\\
     2 & \text{if $r_j > r_i$}
      \end{cases}
\end{equation}

After the partitions are generated, the graph Laplacian will be broken down accordingly. 
With uni-labeling partitioning, the resulted graph Laplacian is $I_N+A$ which is exactly same as the GCN proposed by Kipf et al.\cite{kipf2016variational}.
With distance partitioning and spatial partitioning, the graph Laplacian is dismantled into multiple tensors $A_j$ according to the partition label such that the sum of those tensors equals to $I_N + A$ as follows:
\begin{equation}
\label{partition-sum}
    \matr{I}_N+\matr{A}= \sum_{j}\matr{A}_j.
\end{equation}
Each $A_j$ has its own learnable weight $M$, and the GCN encoder in ST-GCN is calculated as shown in Eqn.~(\ref{st-gcn}) where $\otimes$ denotes the element-wise multiplication. 
\begin{equation}
\label{st-gcn}
    \matr{Z} = \sum_{j} \matr{D}_j^{\frac{1}{2}} (\matr{A}_j \otimes \matr{M}_j) \matr{D}_j^{\frac{1}{2}} \matr{X} \matr{W}. 
\end{equation}
The resulting embedding for the input snapshots are ordered based on their timestamps as the output embedding.

\begin{figure}[ht]
  \centering
  \includegraphics[width=0.90\linewidth]{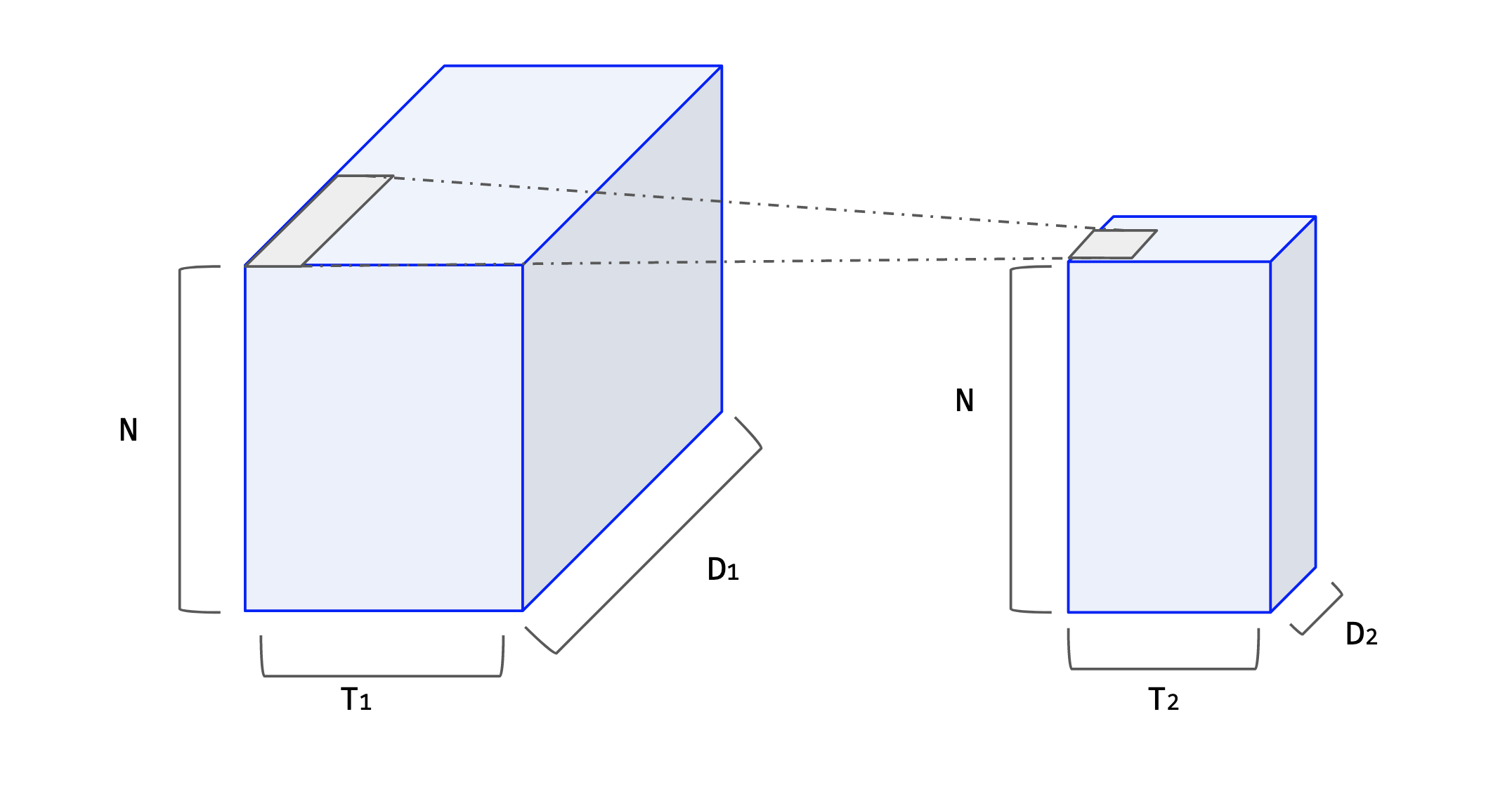}
  \caption{Temporal Convolution}
    \label{t-convolution}
\end{figure}

As shown in Fig.~\ref{t-convolution}, The temporal convolution performs 2-D convolution for each node along its time dimension $T$ and feature dimension $D$ to learn the temporal pattern.

An ST-GCN layers can have multiple temporal and spatial convolutions. 
Multiple ST-GCN layers can be stacked together to construct a deep dynamic graph encoder for better expressive power.
The implementation in Yan et al.\cite{yan2018spatial} ensembles an ST-GCN layer with one spatial convolution in between two temporal convolutions.

Explicit use of time in DTDG has not been explored in the community to the best of our knowledge.
Neither does time prediction task for DTDG. 
It is an interesting direction of future work to develop an explicit time learning model for DTDG and perform time prediction tasks.


\section{Continuous Time Dynamic Graph Model}
\label{sec-CTDG}
DTDG is a well-explored dynamic graph model, which offers plenty of learning algorithms for downstream applications.
However, as discussed above, it is possible to lose important temporal information when DTDG storage models are applied with inappropriate observation frequency.
To preserve all temporal information, we can use \emph{Continuous Time Dynamic Graph(CTDG)} storage model.
CTDG is also called streaming graph.
Under this storage model, the dynamic graph is modeled and stored as the graph updating event stream.
Because all the changes and their timestamps are kept in the database, there is no loss of temporal information.

\begin{defn} 
\emph{Continuous Time Dynamic Graph (CTDG)}: \emph{a dynamic graph $G_{T}$ with $T=[t_0:t_n]$ is regarded as a CTDG model if it is stored as $G_{T} = (G_{t_0}, O_{[t_1:t_n]})$ with $O_{[t_1: t_n]}$ to be a collection of timestamped graph updating events observed during the time span $[t_1:t_n]$,  $G_{t_0}$ to be its initial state at $t_0$. 
Each event $o_{t_i} \in O_{[t_1:t_n]}$ could be either a node updating event $x_{v_i,t_i}$ or edge updating event $e_{\{i,j\},t_j}$.}
\end{defn}

CTDG learning algorithm aims to learn the network evolution embedded in the events stream.
As shown in Fig.~\ref{ctdg-learning}, the framework for CTDG learning algorithms is very similar to that of the dynamic DTDG encoder (the simple decoder framework in Fig.~\ref{DTDG-dynamic}).
In this framework, the input is the observed updating event stream, and then the dynamic CTDG encoder transforms the input event stream to a node-wise graph embedding that learns the graph's temporal pattern in its evolution. 
The encoder learns the regular temporal pattern as described by Eqn.~(\ref{e-pattern}).
Hence it can not only be used directly with common decoders such as MLP\cite{gardner1998artificial} and SVM\cite{suykens1999least}, but also be used with sequential decoders introduced in Sec.~\ref{sec-DTDG}.
Moreover, this framework is commonly applied to analyze interaction networks\cite{trivedi2017know,chang2020continuous}, transaction networks\cite{wang2021apan}, and knowledge graphs in recommendation systems\cite{wang2021apan}.

There are three major challenges in the supervised learning of CTDG: \emph{Event Expiration}, \emph{Computational Exhaustive in Adjacency Matrix Retrieval}, and \emph{Temporal Information Learning}.

\subsubsection{Event Expiration} This is the staleness problem proposed by Kazemi et al.\cite{kazemi2020representation} for CTDG learning. 
How can we determine if long-ago updating events have large impacts on the current nodes?
For example, the relations between users in a social network are defined by their phone call activities, and a phone number is previously abandoned and is recently assigned to a new user. 
In this case, the previous events for the node represented by this number should be expired and no longer reflect its current owner's social relationship, and this node should be counted as a new user without history. 

\subsubsection{Computational Exhaustive in Adjacency Matrix Retrieval} CTDG is stored as a collection of updating events. To obtain its adjacency matrix, we need to scan the whole history and construct the relation between nodes to fill in the cells in the adjacency matrix for each source and destination node pair, according to the observed events and their timestamp. 
However, the computing resources are costly, so that we try to avoid the adjacency matrix construction in CTDG learning.

\subsubsection{Temporal information Learning} 
Each observation in CTDG has its own timestamp and hence rich temporal information can be learnt by analyzing their timestamps. 
The challenge of learning the temporal information is how we can properly use the timestamps as features to learn from. 

In the following sections, we will review different dynamic graph encoders for CTDG and discuss how these graph encoders tackle the aforementioned challenges. 
All summarized CTDG encoders are listed in Table \ref{tab:encoders-ctdg}:
	\begin{table*}
		\centering
		\caption{CTDG Graph Encoders}
		\label{tab:encoders-ctdg}
		\begin{tabular}{c|l|l} 
		\hline
			Time Model & Graph type & Encoder \\
			\hline
		  Implicit & Non-attributed Dynamic Graph & Temporal Random Walk-Based Encoder\cite{nguyen2018continuous,de2018combining,bastas2019evolve2vec,trivedi2019dyrep}\\\hline
			
		  Explicit & Attributed Dynamic Graph & Temporal Attention Based Encoder \cite{xu2020inductive,Xu2019}\\
			  \cline{3-3}
			& &RNN based Encoder\cite{rossi2020temporal,kumar2018learning, kumar2019predicting,trivedi2017know,trivedi2019dyrep,wang2021apan}\\\hline
		\end{tabular}
	\end{table*}

\subsection{Implicit Time CTDG Encoder}
The temporal process regarding to the graph topology can be learnt through the \textit{Temporal Random Walk-Based Encoder}.
Temporal Random Walk is defined as a random walk performed respecting to the timestamp of the edge updating events\cite{masuda2020guide, holme2005network, kempe2002connectivity, berman1996vulnerability}. Once the sets of temporal walks are sampled, random walk-based static graph encoders can be used to generate the embedding. Nguyen et al.\cite{nguyen2018continuous} propose a temporal random walk-based encoder for non-attributed dynamic graphs with three strategies to select the next hop in a walk. De Winter et al.\cite{de2018combining} and Bastas et al.\cite{bastas2019evolve2vec} convert the CTDG to DTDG and perform temporal random walks.
Since random walk-based approaches are generally applied in non-attributed dynamic graphs, practitioners have to combine them with a decoder that utilizes the graph attributes for attributed dynamic network. 
One potential direction of future works can extend temporal random walk-based encoders for attributed dynamic graph by biasing the hop selection based on TPP.
The temporal attentive aggregation for neighborhood message passing in DyREP\cite{trivedi2019dyrep} is an example, in which the maximum walk length is 1 and the probability of next hop is calculated based on TPP.

\subsection{Explicit Time CTDG Encoder}
CTDG is stored as a stream of observed updating events $O(T)$ which could be viewed as sequential data.
Therefore sequential learning models are naturally used in dynmaic CTDG encoder to transform a given node's updating events to its embedding.
One of such kind of encoders is the Temporal Attention based encoder.

\subsubsection{Temporal Attention Based CTDG Encoder}
Inspired by the success of network attention mechanism in learning sequence data\cite{vaswani2017attention}, Xu et al. proposed the \emph{Temporal Attention} mechanism as the dynamic graph encoder in Temporal Graph Attention Network (TGAT). 
It assumes that more critical temporal information are revealed in the relative time span, compared to the absolute time value.
However attributes self-updating is not considered in its temporal learning framework.

With this assumption, Temporal Attention applies Time2Vec\cite{kazemi2019time2vec} to capture critical temporal information from the dynamic graph.
Time2Vec aims to generate a simple vector representation of time so as to enable different learning algorithms to learn the temporal correlation as well as the periodicity with the explicit use of time.
Given a scalar notion of time $\Delta t$, which could be time difference in CTDG setting, its naive Time2Vec encoding $\text{t2v}(\Delta t)$, is a vector of predefined size $d$ and calculated as:
\begin{equation}
    \label{time2vec}
    \text{t2v}(\Delta t)[i] =    \begin{cases}
     w_i \Delta t + \varphi_i & \text{if $i = 1$},\\
     f(w_i \Delta t + \varphi_i) & \text{if $1 < i \leq d$},
      \end{cases}
\end{equation}
where $w_i$ and $\varphi_i$ are trainable weight or predefined weight, and $f$ is a periodic function, such as $\text{sin}(\cdot)$ and $\text{cos}(\cdot)$. 
Time2Vec helps the learning model to learn temporal correlation by simple linear time projection when $i=1$; and it helps to learn the periodicity of time by kernel learning  when $1 < i \leq d$ \cite{xu2020inductive}. 

TGAT applied a modified version of Time2Vec to calculate the functional encoding of a given node $v_i$ at time $t$:
\begin{subequations}
    \begin{align}
   \vecr{h}_{v_i,v_j} &=  \vecr{x}_{v_j,t_j}||\vecr{e}_{ij,t_j}||\text{t2v}(t_i - t_j)
   \label{self-encoding}\\
   \begin{split}
   \matr{H}_{v_i} &= [\vecr{h}_{v_i,v_i}, \vecr{h}_{v_i,v_1},\cdots, \vecr{h}_{v_i,v_j}, \cdots,\vecr{h}_{v_i,v_n}]^{\textbf{T}}\\ &\forall v_j \in N(v_i)  
   \label{functional-encoding}
   \end{split}
   \\
   \begin{split}
   \text{t2v}(\Delta t) &= \sqrt{\frac{1}{d}}[\text{cos}(w_1\Delta t),\text{sin}(w_1\Delta t),\cdots,\\
   & \text{cos}(w_d\Delta t),\text{sin}(w_d\Delta t)]
\label{temporal-kernel}
\end{split}
    \end{align}
\end{subequations}
where $\vecr{x}_{v_i,t_i}$ is the node feature vector for node $v_i$ at time $t_i$;  $\vecr{e}_{ij,t_j}$ is the edge feature vector for the edge between node $v_i$ and $v_j$ at time $t_j$.
In Eqn.~(\ref{self-encoding}), if $v_j = v_i$, $\vecr{e}_{ij,t_j}$ is vector of zeros.

With the functional encoding, temporal attention calculates the query, key and value for $v_i$ at time $t_i$ as:

\begin{equation}
    \begin{split}
    \matr{Q}_{v_i,t_i} &= [\matr{H}_{v_i}]_{0} \matr{W}_Q,\\
    \matr{K}_{v_i,t_i} &= [\matr{H}_{v_i}]_{1:n} \matr{W}_K,\\ 
    \matr{V}_{v_i,t_i} &= [\matr{H}_{v_i}]_{1:n} \matr{W}_V, 
    \end{split}
    \label{functional-attention}
\end{equation}
and the output state of the target node $v_i$ is given by Eqn.~(\ref{general-attention}): 
\begin{equation}
\label{tgat}
   \vecr{z}_{v_i,t_i} = \text{attn}(\matr{Q}_{v_i,t_i}, \matr{K}_{v_i,t_i},\matr{V}_{v_i,t_i}).   
\end{equation}

\subsubsection{RNN-based CTDG Encoder}
As discussed in Sec.~\ref{time-model}, external factors have major contribution to a dynamic graph's evolution.
\emph{RNN-based CTDG encoder} follows the \threestage to capture the external factors contributing to the node attribute self-update. 
RNN-based encoder first generates impression from the observed events related to the target node by a memory function,  and then maintains its memory related to the target node by feeding the impression to a sequential model.
Node embedding is generated from the maintained memory optionally together with other factors, such as node embedding from other kinds of encoder, embedding from the observed events, and embedding from the timestamps.

\emph{Temporal Graph Neural Network (TGN)\cite{rossi2020temporal}} provides a general framework for RNN based encoder.
\emph{TGN} consists of two components which are the memory component and embedding component. 
The memory component represents the model's memory for a given node's history. 
we can denote the memory of node $v_i$ at time $t$  by vector $\vecr{s}_{v_i,t}$.
$\vecr{s}_{v_i,t}$ is updated when an updating event involving node $v_i$ is encountered.
If the event is a node-wise event with new node attribute vector $\vecr{x}_{v_i,t}$, the message to $\vecr{s}_{v_i,t}$ will be calculated as shown in Eqn.~(\ref{tgn-node-message}), where $t^-$ is the timestamp for the last observation for node $v_i$ before $t$ and $\Delta t$ is the time span between $t^-$ and $t$:
\begin{equation}
         \vecr{m}_{v_i,t} = \text{msg}_n(\vecr{s}_{v_i,t^-}, \Delta t, \vecr{x}_{v_i,t})
         \label{tgn-node-message}
\end{equation}

For an interaction event with new edge feature vector $\vecr{e}_{ij,t}$ indicating node $v_i$ as source and $v_j$ as destination, the message is calculated as shown in Eqn.~(\ref{tgn-edge-message-i}) and (\ref{tgn-edge-message-j}), respectively. 
\begin{subequations}
    \begin{align}
         \vecr{m}_{v_i,t} &= \text{msg}_s(\vecr{s}_{v_i,t^-},\vecr{s}_{v_j,t^-}, \Delta t, \vecr{e}_{ij,t},\label{tgn-edge-message-i}\\
         \vecr{m}_{v_j,t} &= \text{msg}_d(\vecr{s}_{v_j,t^-},\vecr{s}_{v_i,t^-}, \Delta t, \vecr{e}_{ij,t}).
         \label{tgn-edge-message-j}
    \end{align}
\end{subequations}
For an undirected network, $\text{msg}_s$ and $\text{msg}_d$ share the same parameters.  

Once the messages for all input events with the target node $v_i$ involved are generated, they are aggregated as described in Eqn.~(\ref{tgn-memory}), where $t^- <t_1, \dots , t_b <= t$:
\begin{equation}
\label{tgn-memory}
      \vecr{s}_{v_i,t} = \text{mem}(\vecr{s}_{v_i,t^-}, \text{agg}(\vecr{m}_{v_i,t_1}, \dots, \vecr{m}_{v_i,t_b} ))
\end{equation}
The $\text{agg}(\cdot)$ refers to a aggregation function, which can be a trainable deep learning layer (e.g., RNN, LSTM), or a simple aggregation without trainable weight (e.g., the mean of those messages, the most recent message).
The $\text{mem}(\cdot)$ function is a deep learning layer representing the memory of the model, and should be selected from the RNN family. 
The output of the memory component $s_{i,t}$ can be used directly as the resulted node embedding as in Jodie\cite{kumar2018learning, kumar2019predicting}, Know-Evolve\cite{trivedi2017know} and DyREP\cite{trivedi2019dyrep}.

However, as proposed by Kazemi et al. \cite{kazemi2020representation}, there is a so-called memory staleness problem for nodes that are not active for a relatively long time depending on the context.
The memory $\vecr{s}_{v_i,t^-}$ of these kind of node is not updated for a long time.
Once a new event arrives, the outdated memory has the same impact as a recent memory.
When calculating the new memory $\vecr{s}_{v_i,t}$,  this is not desirable when the events happened long time ago has not much impact to the graph's evolution.
For example, two users with past frequent connections may not be currently connected in a call social network, since one of the users changes the phone number. 
A new event for an abandoned number represents that this number is recycled and used by a different user, and the history for this number is no longer meaningful.
To overcome this challenge, TGN proposed the embedding component which uses a time embedding \cite{kazemi2019time2vec, xu2020inductive} as one of the input features in inference and could be trained to recognize the impact of staled memory.

The embedding component $\vecr{z}_{v_i,t}$ has a general form as the following message passing schema:
\begin{equation}
    \vecr{z}_{v_i,t} =\sum_{v_j\in N_t(v_i)} f(\vecr{s}_{v_i,t},\vecr{s}_{v_j,t_j},e_{ij,t_j},\vecr{x}_{v_i,t},\vecr{x}_{v_j,t}),
    \label{tgn-embeding}
\end{equation}
where $f(\cdot)$ is a neighbour node aggregation function and $t_j$ is the timestamp that the last edge updating event observed between $v_i$ and $v_j$.

The embedding component $\vecr{z}_{v_i,t}$ has different implementations and can generate embedding without neighborhood information aggregation.
For example, it could be as simple as just the identify function of the memory $\vecr{s}_{v_i,t}$; the time projection used in Jodie \cite{kumar2019predicting}:
\begin{equation}
    \label{time-projection}
    \vecr{z}_{v_i,t} = (1 + \Delta t w) \cdot \vecr{s}_{v_i,t}.
\end{equation}

A MLP-TGAT  network structure takes the states of the target node $v_i$ and its neighbors from the last layer and the Time2Vec embedding of their last updating events' time stamps $\text{t2v} (t-\vecr{T}_{N(v_i)})$ as input, the input state $\vecr{h}_{v_i,t}^{(1)}$ of the first layer is the memory $\vecr{s}_{v_i,t}$ and the output state $\vecr{h}_{v_i,t}^{(l)}$ of the last layer would be the resulting embedding:
\begin{equation}
    \begin{split}
    \vecr{h}_{v_i,t}^{(1)} &= \vecr{s}_{v_i,t}\\
    \vecr{\hat{h}}_{v_i,t}^{(l)} &= \text{TGAT}(\vecr{h}_{v_i,t}^{(l-1)},\text{t2v}(0), \matr{H}_{N(v_i)}^{(l-1)},\text{t2v}(t-\vecr{T}_{N(v_i)}) )\\
    \vecr{h}_{v_i,t}^{(l)} &= \text{MLP}(\vecr{h}_{v_i,t}^{(l-1)}, \vecr{\hat{h}}_{v_i,t}^{(l)})
    \end{split}
\label{tgn-tgat}
\end{equation}

Similarly, Temporal Graph Sum is proposed in TGN \cite{rossi2020temporal}:

\begin{equation}
    \begin{split}
      \vecr{\hat{h}}_{v_i,t}^{(l)} &= \text{ReLU}(\sum_{v_j\in N(v_i)}\matr{W}_1^{(l)}(\vecr{h}_{v_j,t_j}^{(l-1)},\text{t2v}(t-t_j) )),\\
    \vecr{h}_{v_i}^{(l)} &= \matr{W}_2^{(l)}(\vecr{h}_{v_i,t}^{(l-1)}|| \vecr{\hat{h}}_{v_i,t}^{(l)}).
    \end{split}
       \label{tgn-tgs}
\end{equation}

APAN\cite{wang2021apan} applies an asynchronous mail propagator to over come the out-of-order event arrivial issue in TGN's node memory units design.
In on-line training, graph updating events are not guaranteed to arrive in timestamp order.
This will bring instability to RNN based CTDG encoder such as TGN. 
The asynchronous mail propagator fixes this issue by storing the incoming events in their timestamp order.

\subsection{Explicit Time CTDG Decoder}
As discussed, the decoder in CTDG learning can be a simple supervised classifier, such as MLP\cite{gardner1998artificial} and SVM\cite{suykens1999least}.
The application of sequential decoders is introduced in Sec.~\ref{sec-DTDG}.
Temporal Point Process Based Decoder in CTDG learning is introduced in this section.

\paragraph{Temporal Point Process Based Decoder}
Know-Evolve\cite{trivedi2017know} and TDIG-MPNN\cite{chang2020continuous} applies TPP as decoder in edge focus tasks and time predicting tasks with CTDG setting.

Instead of modeling the set of observations $O_{[0:t]}$ as sequential data, temporal point process models $O_{[0:t]}$ as a random process with parameters the input time $t$ and the observations $O_{[0:t^-]}$ before $t$\cite{rasmussen2011temporal}:
\begin{equation}
    \label{tpp}
    f(t,O_{[0:t^-]}) = \lambda(t,O_{[0:t^-]}) \text{exp}(-\int_{t^-}^{t} \lambda(t,O_{[0:t^-]}) dt ),
\end{equation}
where $f(t,O_{[0:t^-]})$ is the probability density function representing an event occurs at time $t$ given the previous observations.
$t^-$ is the time for the last observation right before $t$.
$\lambda(\cdot)$ is the conditional intensity function.
Its form depends on the temporal pattern to capture, such as Poisson process, Hawkes process\cite{hawkes1971spectra}, Self-correcting process\cite{isham1979self}, Power Law and Rayleigh process\cite{miller1958generalized}.
In Know-Evolve, the conditional intensity function $\lambda(\cdot)$ for the target edge $e_{i,j}$ at time $t$ is described as follows:
\begin{equation}
    \label{tpp-decoder}
    \lambda_{i,j}(t|t^-) = \text{exp}(\vecr{z}_{v_i,t^-}^{\textbf{T}} \cdot \matr{W}_{i,j} \cdot \vecr{z}_{v_j,t^-}) \dot (t-t^-),
\end{equation}
where $\matr{W}_{i,j} \in \mathcal{R}^{d \times d}$ is the unique trainable parameter regarding the edge $e_{i,j}$.
For a Rayleigh process, the survival term in Eqn.~(\ref{tpp}) is calculated as: 
\begin{equation}
    \label{tpp-decode-survival}
    \text{exp}(-\int_{t^-}^{t} \lambda_{i,j}(t|t^-) dt )= \lambda_{i,j}(t|t^-) \cdot (t^2-(t^-)^2).
\end{equation}

In the inference, practitioners can estimate the probability that an event happens for the given edge in the next moment, by the product of the resulted probability density and the time duration in interest to perform edge focus task and time predicting task. 
Graph Hawkes Neural Network applies Hawkes process in its decoder\cite{han2020graph} in a similar manner.
It would be interesting for a future work to extend TPP based decoder for node focus task as well as graph focus task.
Also, applying TPP based decoder in DTDG learning for time predicting task is an interesting future work as well.



\section{Challenges And Future Works}
\label{sec-fu}
In this section, we will highlight some challenges and interesting future works in the supervised learning for both DTDG and CTDG.

\subsubsection{Explicit Time DTDG Learning}
DTDG is a very  well-explored dynamic graph model. 
Because its snapshot based representation fits naturally with static graph encoder,
most of the works focus on performing node focusing, edge focusing and graph focusing tasks with the framework of static graph encoder and sequential decoder.
However, since time is implicitly learnt in the decoder with this framework, it is not capable to perform time predicting tasks.
To the best of our knowledge, time predicting task is never considered with DTDG.
Explicit time prediction in DTDG learning will be an important research direction in our opinion.
For example, in financial stock price prediction, we can model each company in the investment portfolio as a node based on the background of the companies, and then the relation between nodes can be formulated as a graph.
In addition to predicting the future price of a particular company's stock, investors are also interested in knowing the time required to hold the stock until the targeted share price is reached.
To facilitate explicit time prediction in both supervised and unsupervised DTDG learning, an explicit time DTDG encoder or decoder is required.
Future works could be conducted on developing and applying explicit time supervised DTDG learning algorithms to predict the number of snapshots required for a particular change to happen.

\subsubsection{Large Scale Dynamic Graph Learning}
GNN based learning models suffer from high memory and CPU power requirement, which becomes more critical in large scale dynamic graph learning. 
We observe a recent trend in solving this challenge in static graph learning\cite{jia2020improving,bai2021efficient,gallicchio2020fast}, while there is still large improvement room to address this issue in dynamic graph learning.
The optimization for existing dynamic graph learning methods to accommodate large scale dynamic graph is a meaningful and interesting future work.  

\subsubsection{The Significance of Temporal Pattern}
As discussed in Sec.~\ref{time-model}, the evolution of dynamic network can be described by Eqn (\ref{e-pattern}), where the future state of a dynamic graph is correlated to its past. 
The ultimate goal of dynamic graph learning is to estimate this function and apply it in different graph learning tasks.
What if the evolution is just a random event that is not correlated with a graph's history?
Or more possibly, there are some randomness in the evolution that cannot be explained or learnt.
In this case, we believe the performance gain by applying dynamic graph learning methods over static graph learning methods depends on how much randomness presented in the temporal process.
Considering the additional complexity of dynamic graph modeling and learning over static ones, the performance gain via evaluating the significance of temporal pattern in a dynamic graph is very useful for industry practitioners to choose an appropriate model with low cost. 

\subsubsection{Implementation Tools for Dynamic Graph Learning}
There are Tensorflow\cite{abadi2016tensorflow}, Pytorch\cite{paszke2019pytorch} and Scikit-Learn\cite{pedregosa2011scikit} to help write data pipelines in machine learning algorithm implementations. Some necessary and useful tools, such as DGL\cite{wang2019deep}, Spektral\cite{grattarola2020graph}, can help implement static graph learning algorithms.  
However, there are less library supports for implementing dynamic graph learning. 
A future work of creating such convenient tools will be beneficial for both the industry as well as the research community.


\section{Conclusion}
\label{conclusion}
As a data structure, graph data modeling attracts the data science community by its extraordinary expressive power. 
The world is a dynamic system evolving along time, and so does the data generated from real world activity. 
Dynamic graph is the extension of naive graph modeling to capture this temporal dynamic.
Moreover, with the recent development of big data technology and online services, recording and storing graph attributes become feasible. 
This urges the need for effective supervised dynamic graph algorithms to perform different machine learning tasks with better accuracy.
To provide a comprehensive reference for academia researchers and industry practitioners, this survey is conducted on the following scopes:
\begin{itemize}
    \item Background of dynamic graph learning with a full-scale systematic summary from storage model, learning purpose, algorithm architecture framework. 
    \item We propose the \threestage. Based on the proposed temporal learning framework, we discuss how temporal information is learnt by different dynamic graph learning algorithms. \threestage~also provides a general mathematical form of dynamic graph learning algorithms 
    As far as we know, this is the first and only temporal learning framework which could be applied to both attributed and non-attributed dynamic graph learning.
    \item Supervised dynamic graph learning algorithms with detailed introduction from DTDG compatible algorithms to CTDG compatible algorithms, from encoders to decoders and from implicit time algorithms to explicit time algorithms. 
    \item Future research directions according to the topics discussed
\end{itemize}

We hope that with this survey paper, we can offer a convenient reference to industry practitioners and facilitate future research for the academic community.

\bibliographystyle{./bibliography/IEEEtran}
\bibliography{./bibliography/main}
\begin{appendices}

	\begin{table*}
		\centering
		\caption{List Of Notations}
		\label{tab:notation}
		\begin{tabular}{ll} 
		Notation & Description \\
		\hline
		$\alpha$ & The smoothing factor of EMA\\ 
		$G$ & A graph \\ 
		$V$ & Set of nodes $V={v_1, v_2, \cdots v_n}$; the input value $\matr{V}$ of the attention layer\\
		$V(T)$ & Node-wise events observed at period $T$\\
		$E$ & Set of edges $E={e_1, e_2, \cdots e_n}$\\
		$E(T)$& Interaction events observed at period $T$\\
		$v$ & A node $v$ in $V$\\
		$e$ & An edge $e$ in $E$\\
		$e_{i,j}$ & the edge start from node $i$ to $j$\\ 
		$\vecr{e}_{ij,t}$ & the edge feature vector for edge $e_{i,j}$ at time $t$\\
		$t$ & Time step / event time\\
		$t^-$ & Time step just before time $t$ \\
		$T$ & Time duration\\
		$G_T$ & Dynamic graph in time duration $T$\\
		$O_T$ & Observations of a dynamic graph in time duration $T$\\
		$G_t$ & Dynamic graph at time $t$\\
		$o_t$ & Observation of a dynamic graph at time $t$, $o_t \in O_T$\\
		$o_t(v_{i})$ & A node updating event for node $v_i$ observed at time $t$\\
		$e_{\{i,j\},t})$ & An edge updating event between node $v_i$ and $v_j$ observed at time $t$\\
		$Y$ & Set of labels in a data set\\
		$y_i$ & A particular label $y_i$ in $Y$\\
		$\matr{X}$ & Nodes feature matrix for $V$ \\
		$\vecr{x}$ & Node feature vector in $X$ \\
		$w, b, j, \varphi$ & Learnable or preset model parameters in scala form\\
		$\vecr{w, b, j, \varphi}$ & Learnable or preset model parameters in vector form\\
		$\matr{W}, \matr{B}, \matr{J}$ & Learnable or preset model parameters in matrix form\\
		$f,\lambda$ & Model functions\\
		$h,c$ & Hidden states in scala form\\
		$\vecr{h,c}$ & Hidden states in vector form\\
		$\matr{H},\matr{C}$ & Hidden states in matrix form\\
		$\vecr{h}_{v_i}^k$ & Hidden state of node $v_i$ at layer $k$ in a learning model\\
		$N(v_i)$ & Neighborhood function which returns the neighboring nodes for the input node $v_i$\\
		$E(v_i)$ & A function returns all edges connecting $v_i$ to its neighbors\\
		$\matr{N_t}$ & Neighborhood aggregation matrix, formed by stacking each node's neighboring aggregation\\
		$\matr{Q}$ & The input query of the attention layer\\
		$\matr{K}$ & The input key of the attention layer; Kalman gain in Kalman filter\\
		$\matr{P}$ & Positional encoding in TSA; Covariance matrix in Kalman filter\\
		$\matr{K^{\textbf{T}}}$ & The superscript $T$ in bold text represents the transpose operation in matrix\\
		$\vecr{z}_{v_i}$ & Embedding vector for node $v_i$\\
		$\matr{Z}$ & Graph embedding matrix\\
		
		$\matr{A}$ & Adjacency matrix\\
		$\matr{\hat{D}}$ & Diagonal degree matrix\\
		$\matr{L}$ & Graph labracian\\
		$\vecr{m}_{v_i,t}$ & Message passed to node $v_i$ at time $t$\\
		$\text{out}(\cdot)$ & output function of a learning model\\
		$\text{tp}(\cdot)$ & Temporal Pattern to learn in \threestage \\
		$\text{asu}(\cdot)$ & Attributes Self-Updating function in \threestage\\
		$\text{ap}(\cdot)$ & Association Process function \threestage\\
		$\text{mp}(\cdot)$ & Message Passing function \threestage\\
		$\text{attn}(\cdot)$ & Attention layer\\
		$\text{msg}(\cdot)$ & message generating function in TGN\\
		$\text{mem}(\cdot)$ & The memory of a TGN model\\
		$\text{agg}(\cdot)$ & message aggregating function in TGN\\
		$\text{t2v}(\cdot)$ & Time2Vec embedding \\
		$\otimes$ & Element-wise multiplication\\
		$||$ & Concatenation\\
				\hline
		\end{tabular}
	\end{table*}
	
		\begin{table*}
		\centering
		\caption{List Of Abbreviations}
		\label{tab:abbreviation}
		\begin{tabular}{ll} 
		Abbreviation & Description \\
		\hline
		APAN & Asynchronous Propagation Attention Network\\
		CNN & Convolutional Neural Network\\
		Conv1d & 1-d Convolution Neural Network\\ 
		CTDG & Continuous Time Dynamic Graph\\
		DeepWalk & A graph embedding method based on deep learning and random walk\\
		DNGR & Deep Neural Graph Representation\\
		DTDG & Discrete Time Dynamic Graph\\
		DyGGNN & Dynamic Gated Graph Neural Network\\
		DySAT & Dynamic Self-Attention Network\\
		EvolveGCN & Evolving GCN\\
		EGCU & Evolving GCN unit\\
		EMA & Exponential Moving Average\\
		ARIMA & Auto-Regressive Integrated Moving Average\\
		FNN & Feedforward Neural Network\\
		GCN & Graph Convolution Neural Network\\
		GNN & Graph Neural Network\\
		GraRep & Graph representation with global structural information\\
		GRU & Gated Recurrent Unit\\
		HARP & Hierarchical Representation Learning\\
		HOPE & High Order Proximity Preserved Embedding \\
		Jodie & a coupled recurrent neural network model that learns the embedding trajectories of users and items\\
		Know-Evolve & Deep evolutionary knowledge network for dynamic knowledge graph learning\\
		LINE & Large-scale Information Network Embedding\\
		LSTM & Long Short Term Memory\\
		M-NMF & Modularized Non-negative Matrix Factorization \\
		RNN & Recurrent Neural Network \\
		SDNE & Structural Deep Network Embedding\\
		softmax & softmax layer\\
		STGCN & Spatial Temporal Graph Convolution Network\\
		TADW & Text Associated DeepWalk\\
		TGN & Temporal Graph Network\\
		TGAT & Temporal Graph Attention Network\\
		TGNN & Temporal Graph Neural Network\\
		TPP & Temporal Point Process\\
		TSA & Temporal Self-Attention\\
		
		\hline
		\end{tabular}
	\end{table*}

\end{appendices}

\vspace{12pt}

\end{document}